\documentclass[runningheads]{llncs}
\usepackage{graphicx}
\usepackage{makecell}
\usepackage{multirow}
\usepackage{cite}
\usepackage{amssymb}
\usepackage{utfsym}
\usepackage{ragged2e}
\usepackage{amsmath}
\usepackage{booktabs}

\usepackage{colortbl}
\usepackage{xcolor}
\usepackage{float}
\usepackage{authblk}
\usepackage{bm}
\usepackage[hidelinks,colorlinks=true,linkcolor=blue,citecolor=blue]{hyperref}
\newcommand{\ucite}[1]{\cite{#1}}
\begin{document}

\title{Skeleton Supervised Airway Segmentation} 
\titlerunning{Skeleton Supervised Airway Segmentation}
%
%

\author{Mingyue Zhao\inst{1} \and
Han Li\inst{1}\and
Li Fan\inst{2}\and
Shiyuan Liu\inst{2}\and
Xiaolan Qiu\inst{3}\and
S.Kevin Zhou\inst{1}}

%
\institute{Center for Medical Imaging, Robotics, Analytical Computing \& Learning (MIRACLE), School of Biomedical Engineering, University of Science and Technology of China, Suzhou, China \and
Department of Radiology, Second Affiliated Hospital, Naval Medical University, Shanghai, China \and
Suzhou Key Laboratory of Microwave Imaging, Processing and Application Technology Suzhou Aerospace Information Research Institute, Suzhou, China}

\maketitle       

\begin{abstract}
Fully-supervised airway segmentation has accomplished significant triumphs over the years in aiding pre-operative diagnosis and intra-operative navigation. However, full voxel-level annotation constitutes a labor-intensive and time-consuming task, often plagued by issues such as missing branches, branch annotation discontinuity, or erroneous edge delineation. label-efficient solutions for airway extraction are rarely explored yet primarily demanding in medical practice. To this end, we introduce a novel skeleton-level annotation (SkA) tailored to the airway, which simplifies the annotation workflow while enhancing annotation consistency and accuracy, preserving the complete topology. Furthermore, we propose a skeleton-supervised learning framework to achieve accurate airway segmentation. Firstly, a dual-stream buffer inference is introduced to realize initial label propagation from SkA, avoiding the collapse of direct learning from SkA. Then, we construct a geometry-aware dual-path propagation framework (GDP) to further promote complementary propagation learning, composed of hard geometry-aware propagation learning and soft geometry-aware propagation guidance. Experiments reveal that our proposed framework outperforms the competing methods with SKA, which amounts to only 1.96$\% $ airways, and achieves comparable performance with the baseline model that is fully supervised with 100$\%$ airways, demonstrating its significant potential in achieving label-efficient segmentation for other tubular structures, such as vessels.

\end{abstract}
\section{Introduction}
\begin{figure}[t]
\centering
\includegraphics[width=0.95\textwidth]{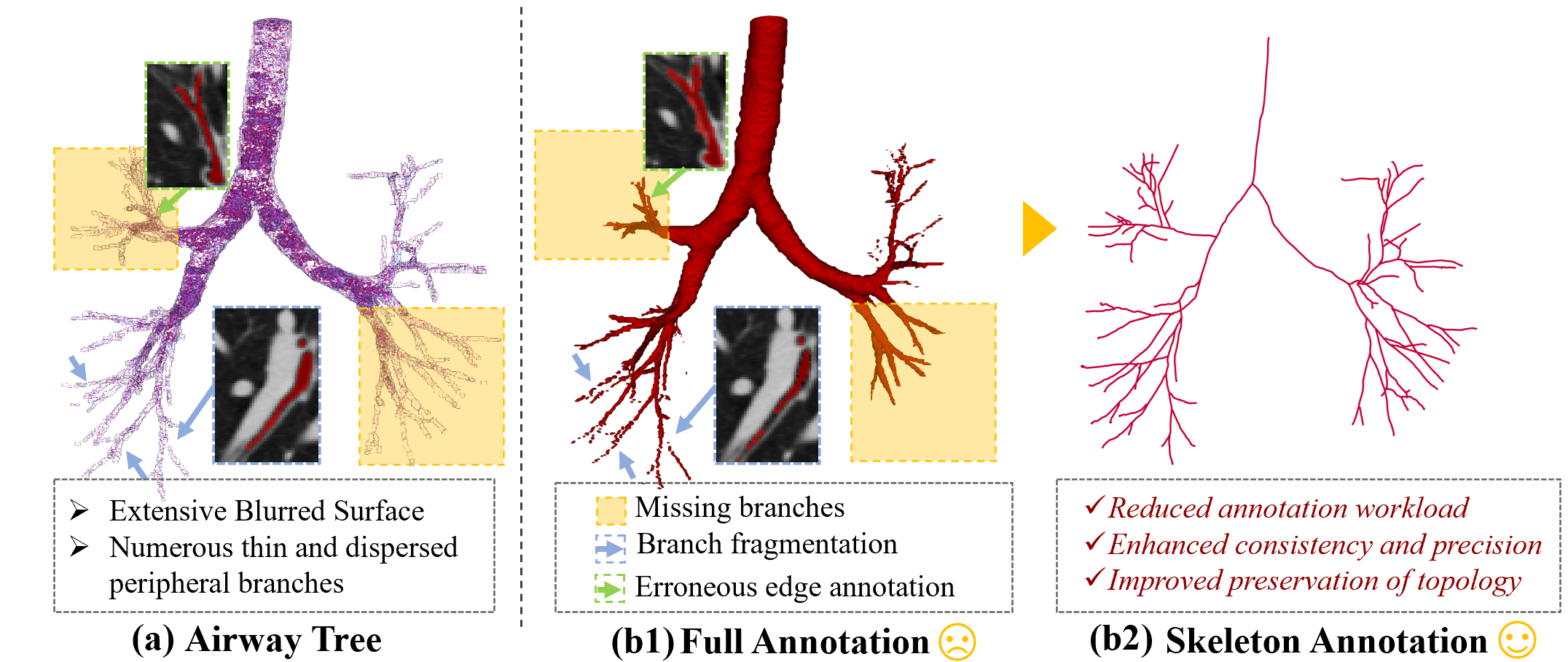}
    \caption{\textbf{Motivation} for introducing the skeleton level annotation for the airway.}
\label{SkA}
\vspace{-5pt}
\end{figure}

Airway segmentation is fundamentally important for early diagnosis, treatment, and ongoing assessment of lung diseases~\ucite{de2020airway,xiao2020prevalence}.  Recently, deep learning-based algorithms~\ucite{qinairwaynetse_2020,qin_Tubule-Sensitive_learning_2021,yu_breakBronchiReconstruction_2022,zheng_alleviating_2021,wang_naviairway_2022,Zheng_Refined_Local_2021,zhang2023towards,zhang2021fda,zhao2019bronchus} have emerged as highly effective tools for extracting airways. Despite this, existing algorithms still face significant challenges in obtaining high-quality voxel-level annotations for the airway, mainly attributed to the inherent characteristics of airway tree: \emph{ 1) Extensive Blurred Surface/Edge Area.} The airway tree necessitates labeling more edge voxels that gradually blur along the depth of the airway. \emph{ 2) Numerous thin and dispersed peripheral branches.} 
These branches occupy a significant proportion and exhibit diverse directions, necessitating annotators to constantly adjust their observation perspectives.
These two factors make voxel-wise annotation a time-consuming and labor-intensive task. Furthermore, compared to the labeling of near-convex structures such as organs or tumors, it inevitably leads to more inconsistent and noisy annotations, such as 
missing branches, branch fragmentation, or erroneous edge annotations (as shown in Fig.~\ref{SkA}). Hence, there is an urgent need to propose 
a label-efficient solution for bronchioles segmentation.

Partial annotation learning~\ucite{zhang2022shapepu,chen2022scribble2d5,lee2021weakly,wei2021scribble,zhou2023weakly,zhang2023partial,xu2021partially,han2023scribble,cai20233d, Wang2023s2me,li2023segment,zhang2022cyclemix,li2022pln,zhou2023progressive} utilizing only a subset of annotated pixels has demonstrated comparable performance to fully supervised learning both in natural images and medical images. It shows great potential in difficult-to-annotate segmentation tasks, yet its application in tubular structures remains unexplored.
Xu~\emph{et al.}~\ucite{xu2021partially} propose a partial supervision manner for vessel segmentation with only a few 2D patches annotated. Zhang~\emph{et al.}~\ucite{zhang2023partial} achieve weakly supervised learning of coronary artery vessel segmentation by randomly selecting partial vessel branches for voxel-level annotation. Despite significantly alleviating the annotation burden compared with the full voxel-level annotation, they still adopt the voxel-wise annotation fashion while paying little attention to topological completeness. The challenge of elaborately delineating 
fine, dispersed branches of the airway tree remains. 
To this end, 
we introduce a novel Skeleton Annotation (SkA) tailored to the airway. 

As illustrated in Fig.~\ref{SkA}, the SkA shows three main superiority over the voxel-level annotations: 1) \textbf{Reduced annotation workload.} SkA, focusing only on the skeleton, is simpler and more efficient than voxel-level labeling, which only needs to annotate 1.96\% of the airway voxels (save 80+\% time) and reduce numerous blurred edge annotations. 2) \textbf{Enhanced annotation consistency and precision.} Voxels annotated in SKA tend to be closer to the lumen center, exhibiting more prominent airway characteristics, which effectively enhances the accuracy and consistency of annotations. 3) \textbf{Improved preservation of topology}. This method is more conducive to maintaining the complete topological structure of the airway, which is vital for accurate modeling and analysis. 

Nevertheless, relying solely on SkA to achieve reliable voxel-wise airway prediction still poses significant challenges: 1) \emph{Extremely sparse supervision}: Since only the skeleton is annotated, the supervision signal is too sparse to support the training of the network. 2) \emph{Limited diversity in supervision signals}: This issue can be observed in two main aspects. The first is the absence of edge annotations, which is crucial for defining precise boundaries in segmentation tasks. Secondly, due to the inherent nature of SkA, this labeling method exhibits a stronger preference~\cite{wang2023blpseg} for annotation positions (i.e., closer to the lumen center) compared to other sparse annotations (e.g., scribbles) that can cover a broader area with an accumulation of extensive annotated samples. This limited annotation bias implies less effective information, leading to impracticality in direct training.

To address this, we further propose a novel skeleton-supervised learning method for airway segmentation. Firstly, it employs a dual-stream buffer inference strategy to propagate knowledge from SkA to unannotated regions, avoiding the collapse of learning directly from skeleton-level annotation. Secondly, considering the tree-like structure, geometry-aware dual-path propagation learning, composed of hard geometry-aware learning and soft geometry-aware guidance, is constructed to achieve accurate airway segmentation.

The main contributions are summarized as follows. \textbf{1)} To the best of our knowledge, we are the first to introduce skeleton-level annotation (SkA) into airway segmentation. It significantly reduces annotation burden while enhancing annotation consistency and preserving more complete topology, aligning more closely with the characteristics of tubular structures.
\textbf{2)} Tailored for airway segmentation, a skeleton-supervised learning method is proposed. Firstly, considering both the spatial locations and grayscale information of SkA, initial label propagation from SkA is conducted utilizing a dual-stream buffer inference strategy (DBI), avoiding the collapse of learning directly from SkA. Secondly, considering the tree-like structure, a geometry-aware dual-path propagation learning method, composed of hard geometry-aware learning and soft geometry-aware guidance, is presented in a dual-supervision manner to further achieve complementary label propagation learning.
\textbf{3)} Extensive experiments indicate that our method achieves comparable performance to the fully-supervised method with only \textbf{1.96\%} annotated voxels. Moreover, our method exhibits greater potential for diverse tubular structure segmentation tasks.
\begin{figure}[t]
\centering
\includegraphics[width=1\textwidth]{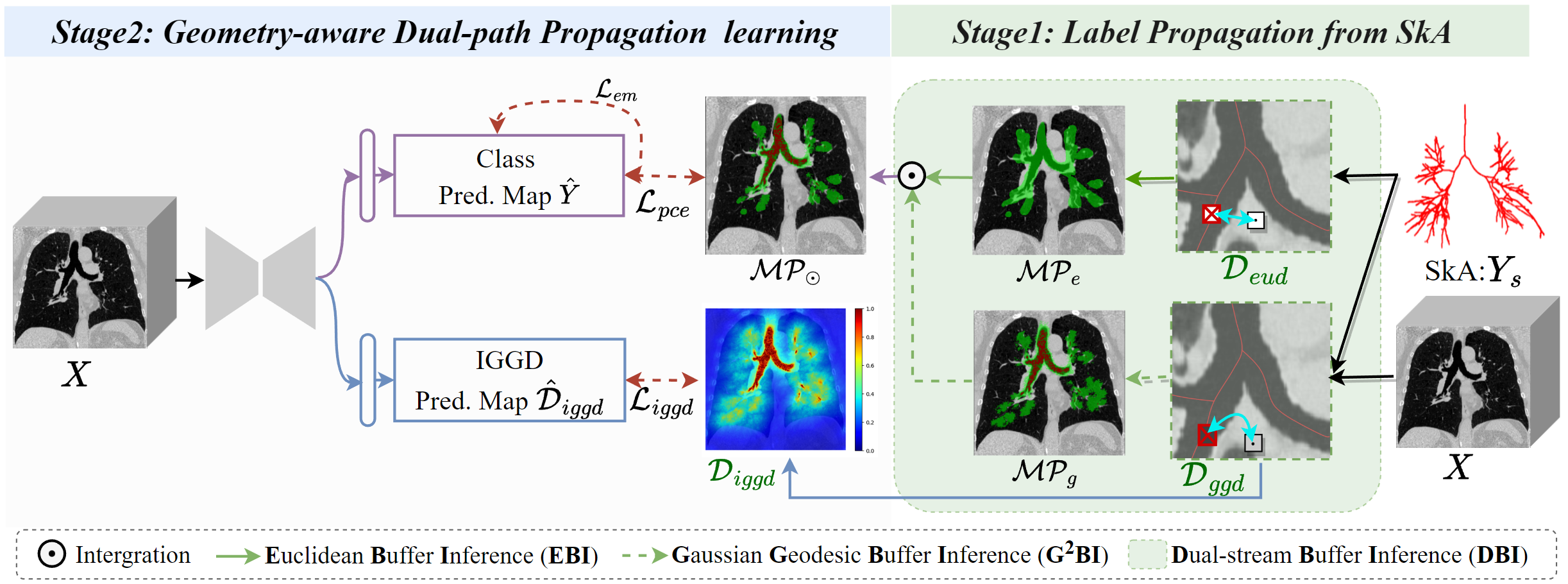}\vspace{-10pt}
\caption{Overview of our skeleton-supervised airway segmentation framework. $\mathcal{MP}_g$, $\mathcal{MP}_e$  and $\mathcal{MP}_\odot$ are the mask proposals after G$^{2}$BI, EBI, and DBI, where \textcolor{red}{red},~\textcolor{green}{green} denotes foreground and unknown regions, the rest area is the background.}
\label{Skeleton-supervised learning}
\vspace{-10pt}
\end{figure}

\section{Methodology} \label{sec:method}

Given a CT volume $X$ with the corresponding SkA $Y_s$, we aim to learn a mapping $\mathcal{F}: X\mapsto \hat{Y}$ based on SkA, where $\hat{Y}$ is the voxel-level airway prediction. As shown in Fig.~\ref{Skeleton-supervised learning}, there are two stages in our methods: 1) Label Propagation from Skeleton Annotations and 2) Geometry-aware Dual-path Propagation Learning. 

\begin{figure}[t]
\centering
\includegraphics[width=1\textwidth]{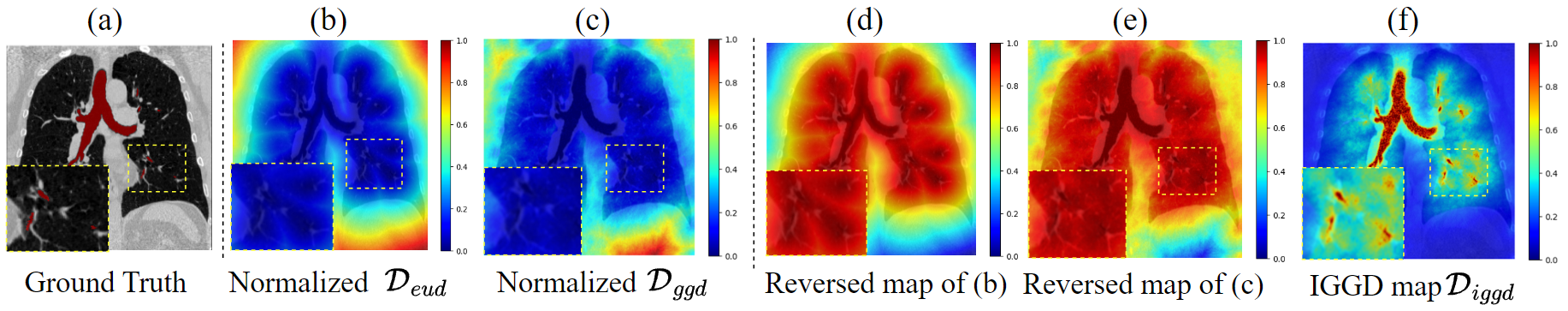}
\vspace{-10pt}
\caption{Visualization of distance maps.  (d) and (e), as the reversed maps of (b) and (c), are used for better visual contrast. The $\mathcal{D}_{iggd}$ (f) is much clearer than (d) and (e). }\label{Distance map}
\end{figure}

\vspace{5pt} 
\noindent \textbf{Label Propagation from Skeleton Annotations.}\label{Label Propagation from Skeleton Annotations }
We propose an effective label propagation method, named Dual-stream Buffer Inference (DBI), to propagate the knowledge from SkA to more unlabeled voxels.  As shown in Fig.~\ref{Skeleton-supervised learning}, it includes two modules:

\underline{Gaussian Geodesic distance Buffer Inference~(G$^{2}$BI).}
Specifically, given a point $i$ in image space $\Omega$, $\Omega_{s} = \left \{ i|Y_{s}(i)=1 \right \} $ denotes the initial annotated voxel set in SkA. 
Geodesic distance~\ucite{wang2018deepigeos} considers the pixels' spatial distance relationship and their appearance similarity, making it better suited to the airway with severe intra-scale imbalance. However, the geodesic distance-based label propagation relies on a strong prerequisite: intra-class pixels have similar grayscale values and show flat gradients. To improve the label consistency within homogeneous areas, Gaussian smoothing is implemented on images before applying the geodesic distance transformation.
Concretely, for each voxel $i \in \Omega$, the Gaussian geodesic distance from $i$ to $\Omega_{s}$ can be formulated as:
\begin{equation}
    \mathcal{D}_{ggd}[i,\Omega_{s},g_\sigma(x)] = \mathop{min}\limits_{j \in \Omega_{s} } \mathcal{D}_{ged} [i,j,g_\sigma(x)], 
\end{equation}
\begin{equation}
    \mathcal{D}_{ged}[i,j,g_\sigma(x)] = \mathop{min}\limits_{p \in \mathcal{P}_{i\to j}, j \in \Omega_{s}}\int_{0}^{1} \left \| \nabla g_\sigma(x) (p(\xi)) \cdot \mathbf{u}(\xi)  \right \| d\xi,
\end{equation}
where $g_\sigma(\cdot) $ denotes a Gaussian filter function with a standard deviation of $\sigma$. $\mathcal{P}_{i\to j}$ is the set of all possible paths from  $i$ to $j$ for $j \in \Omega_{s}$, $p \in \mathcal{P}_{i\to j}$ is one feasible path and parameterized by $\xi \in [0, 1]$. $\mathbf{u}$ is a unit vector tangent to the direction of path $p$. The calculated $\mathcal{D}_{ggd}[i,j,g_\sigma(x)]$ is visualized in Fig.~\ref{Distance map}(c1). 

To facilitate the label expansion of the foreground and background, we exploit bilateral G$^{2}$BI. Let $\Omega_{f_1}$ and $\Omega_{b_1}$ denote the foreground and background after label propagation respectively, then they can be obtained by:
\begin{equation}
    \Omega_{f_1} = \left\{i|i\in \Omega
    \, and\, 
     \mathcal{D}_{ggd}(i)<\delta_{1} max(\mathcal{D}_{ggd})\right\},
\end{equation}
\begin{equation}
    \Omega_{b_1} = \left\{i|i\in \Omega
\, and\,  \mathcal{D}_{ggd}(i)>\delta_{2} max(\mathcal{D}_{ggd})\right\}, 
\end{equation}
where $\delta_{1},\delta_{2} \in [0, 1]$ are hyper-parameters indicating the degree of expansion and $\delta_{1}<\delta_{2}$. The mask proposal $\mathcal{MP}_{g}$ generated from G$^{2}$BI, is depicted in Fig.~\ref{Skeleton-supervised learning}.

\underline{Euclidean distance Buffer Inference (EBI).} The Euclidean distance, as an absolute spatial distance metric, is employed as an additional constraint for label expansion. In light of the pronounced intra-class scale imbalance within the airway tree, where the narrowest branches may measure only one pixel in width, Euclidean distance fails to achieve reliable diffusion of foreground skeleton labels. Consequently, we implement unilateral EBI to achieve background expansion:
\begin{equation}
    \Omega_{b_2} = \left\{i|i\in \Omega-\Omega_{s} 
    \, and\, 
     \mathcal{D}_{eud}(i)>\gamma  max(\mathcal{D}_{eud})\right\}, 
\end{equation}
where $\mathcal{D}_{eud}(i) =\mathop{min}\limits_{j\in \Omega_{s}}  \left \| i -j\right \| _2$ and $\gamma \in [0, 1]$ is a hyper-parameter to control the degree of background expansion. Similarly, $\mathcal{MP}_{e}$, depicted in Fig.~\ref{Skeleton-supervised learning}, denote the generated mask proposal after EBI.

Thus far, the foreground and background after propagation can be represented as $\Omega_{B} = \Omega_{b_1} \cup\Omega_{b_2}$ and $\Omega_{F} = \Omega_{f_1}$, respectively. 
Fig.~\ref{Skeleton-supervised learning} gives an example of final mask proposal $\mathcal{MP}_\odot$. For simplicity, we denote the set of labeled voxels as $\Omega_{L}=\Omega_{B}\cup \Omega_{F}$ and the voxel set with unknown labels as $\Omega_{U}=\Omega-\Omega_{L}$.

\vspace{5pt}
\noindent \textbf{Geometry-aware Dual-path Propagation Learning.}\label{Geometry-Aware Propagation Learning}
As illustrated in Fig.~\ref{Skeleton-supervised learning}, we devise a dual-path network in a multi-head manner to achieve skeleton-supervised learning. The main branch is dedicated to voxel-wise airway prediction, supervised by the proposal masks $\mathcal{MP}_\odot$ (acting as hard geometry-aware information). The Inverse Gaussian Geodesic Distance map (IGGD) serves as fine-grained soft geometry-aware information in the auxiliary branch to aid the training of the main branch. 

\underline{Hard Geometry-Aware Propagation Learning.}
Similarly to ~\ucite{chen2022scribble2d5,zhou2023weakly,wei2021scribble,zhai2023PA-Seg}, we impose the partial cross-entropy loss solely on the annotated pixels ($i\in \Omega_{L}$):
\begin{equation}
    \mathcal{L}_{pce}=-\frac{1}{\left | \Omega _L \right | } \sum_{i\in \Omega _L}^{} (\hat{Y}(i)log(\mathcal{MP}_\odot (i))+(1-\hat{Y}(i))log(1-\mathcal{MP}_\odot (i))).
\end{equation}
Besides, to encourage the model to produce high-confidence predictions for unannotated voxels, entropy minimization~\ucite{grandvalet2004semi} is applied to the class probability prediction maps, with the corresponding supervised loss formulated as:
\begin{equation}
    \mathcal{L}_{em}= -\frac{1}{\left | \Omega _U \right | } \sum_{i\in \Omega _U}^{}\mathcal{MP}_\odot (i)log(\mathcal{MP}_\odot (i)).
\end{equation}

\underline{Soft Geometry-Aware Propagation Guidance.}
Despite that initial label propagation for SkA partially alleviates the problem of sparse supervision, the model still lacks the fine-grained structural perception of the airway.
We find that the Gaussian geodesic distance map~$\mathcal{D}_{ggd}$ implicitly incorporates more fine-grained multiscale-aware and edge-aware information, which is well-suited as geometry-aware information to boost voxel-wise predictions in the main segmentation branch. Considering the similarity in grayscale between lung parenchyma and airways, we perform an inverse transformation on $\mathcal{D}_{ggd}$ to get the IGGD maps: $\mathcal{D}_{iggd}(i)=\mathbf{1}/(\mathcal{D}_{ggd}(i)+c)$, accentuating inter-class difference while avoiding over-segmentation. Where $c=1$ to ensure $\mathcal{D}_{iggd} \in [0, 1]$ while avoiding division by zero. Compared with (e) and (f) in Fig.~\ref{Distance map}, the inverse transformation significantly enhances the airway branches within lung, thereby providing fine-grained and robust auxiliary information for branch segmentation. Mean Squared Error (MSE) loss is adopted in this branch:
\begin{equation}
    \mathcal{L}_{iggd} = \frac{1}{N}{\textstyle \sum_{i=1}^{N}(\hat{\mathcal{D}}_{iggd}-\mathcal{D}_{iggd})^2},
\end{equation}
where $\hat{\mathcal{D}}_{iggd}$ represents the prediction map of the auxiliary branch. Finally, the proposed skeleton-supervised learning framework can be trained by minimizing:
    $\mathcal{L}_{total}=\mathcal{L}_{pce}+ \lambda_1\mathcal{L}_{em}+\lambda_2\mathcal{L}_{iggd}$,
where $\lambda_1$ and $\lambda_2$  are the trade-off weights. Thus far, the mask proposals generated from DBI serve as hard geometry-aware information, providing local absolute supervision guidance for segmentation networks. While the IGGD map as soft geometry-aware information, offers global fine-grained scale-aware and edge-aware guidance for segmentation.

\section{Experiment and Results}\label{experiments}
\noindent{\textbf{Datasets and Evaluation Metrics.}}
Experiments are implemented on the public Binary Airway Segmentation (BAS) Dataset~\cite{qinairwaynetse_2020}, consisting of 90 CT scans. ATM22~\cite{zhang2023multi} dataset consisting of 299 CT scans, is introduced as external validation. SkA of BAS dataset is first extracted using MIMICs software and then manually corrected by a panel of well-trained experts~\ucite{zhao2023gdds}. 
Following~\cite{qin_Tubule-Sensitive_learning_2021,exact_2012,zheng_alleviating_2021, Zheng_Refined_Local_2021}, we adopt volumetric-based metrics (Dice Similarity Coefficient (DSC), True Positive Rate (TPR), and False Positive Rate (FPR)) and topology-based metrics (Branches Detected (BD)~\ucite{exact_2012}, Branches Detected (BD$^*$)~\ucite{qin_Tubule-Sensitive_learning_2021}, Tree-length Detected (TD)~\ucite{exact_2012}) for evaluation. BD and BD$^*$ represent cases where a branch is regarded as detected when 80\% or just one voxel of it is detected. Only the largest component of segmentation results are evaluated and five-fold cross-validation is conducted to obtain the final results. The source code will be available.


\begin{figure}[t]
\centering
\includegraphics[width=\textwidth]{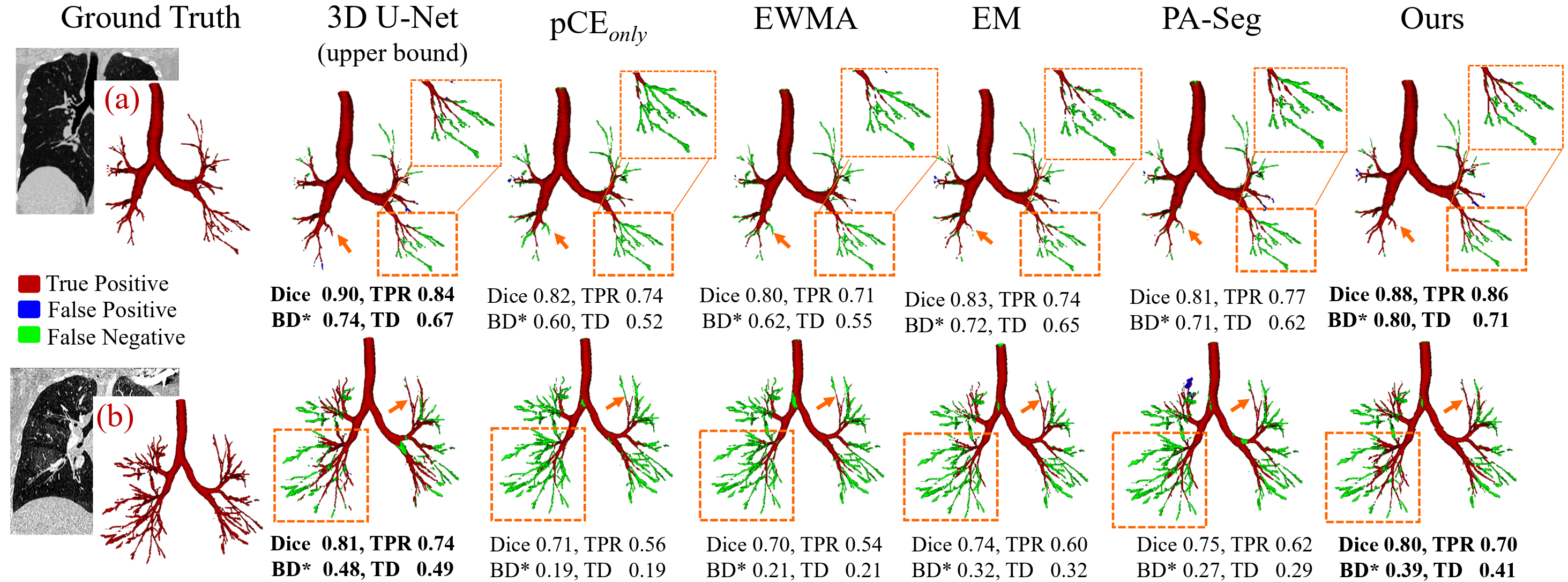}\vspace{-10pt}
\caption{Segmentation results on a moderate case (a) and a challenging case (b). \textcolor{orange}{Orange boxes} and \textcolor{orange}{arrows} highlight the branches with significant segmentation discrepancies.} \label{result}
\end{figure}

\begin{table*} [t]
\centering
    \small
    \caption{Quantitative results based on BAS \cite{qinairwaynetse_2020} and ATM22 dataset~\cite{zhang2023multi}. }
    \label{tab:sota}
    \resizebox{0.95\textwidth}{!}
    {%
        \begin{tabular}{c|c|c|l|c|c|c|c|c|c}
        \toprule
        
          Train Data&Test Data & Annotation & Methods  & DSC(\%) $\uparrow$  & TPR(\%) $\uparrow$ & FPR(\%) $\downarrow$ & BD(\%) $\uparrow$ & BD ${^*}(\%)\uparrow$ & TD(\%)$\uparrow$ \\
        \midrule
        \multirow{10}{*}{BAS}
        &  \multirow{8}{*}{BAS}
        &FA & U-Net~\cite{3dunet}(Upper bound)  & 90.53$_{\pm0.4}$  & 91.13$_{\pm0.7}$  &.032$_{\pm0.007}$&62.75$_{\pm2.4}$& 79.52$_{\pm1.1}$ & 75.10$_{\pm1.6}$ \\
        \cline{3-10}
        &&\multirow{7}{*}{SkA}
        &pCE$_{only}$~\cite{lin2016scribblesup}$^\dag$
        &86.21$_{\pm0.3}$ & 78.99$_{\pm0.7}$ &$\bm{.010_{\pm.004}}$&40.70$_{\pm2.9}$& 60.21$_{\pm3.3}$ &53.06$_{\pm3.4}$\\
        & &&EWMA~\cite{lee2020scribble2label}$^\dag$
        &84.77$_{\pm0.3}$& 77.21$_{\pm0.1}$&.012$_{\pm.001}$& 41.25$_{\pm0.1}$& 61.26$_{\pm0.1}$& 54.48$_{\pm0.2}$  \\
        
        &&&Luo~\emph{et al.}~\cite{luo2022scribble} $^\dag$
        &84.74$_{\pm0.3}$& 77.20$_{\pm0.1}$ &.010$_{\pm.001}$ & 41.18$_{\pm0.1}$& 61.33$_{\pm0.1}$& 54.60$_{\pm0.3}$\\ 
        
        &&&Gatedcrf~\cite{tang2018regularized}$^\dag$
        & 84.92$_{\pm0.7}$& 79.84$_{\pm2.6}$ &.025$_{\pm.010}$ & 45.93$_{\pm2.5}$& 65.35$_{\pm2.3}$& 58.55$_{\pm2.5}$\\ 
        
       & & &PA-Seg~\cite{zhai2023PA-Seg} &85.45$_{\pm0.1}$&81.05$_{\pm0.7}$& .030$_{\pm.005}$  &48.55$_{\pm0.6}$&67.43$_{\pm0.4}$&61.13$_{\pm0.8}$\\ 
        
       & &&EM~\cite{grandvalet2004semi}$^\dag$
        &${87.62_{\pm0.6}}$& 81.20$_{\pm0.7}$ &$\bm{.010_{\pm.001}}$ & 51.64$_{\pm1.3}$& 69.69$_{\pm1.2}$& 63.93$_{\pm1.4}$
        \\ 
        
       & && \textbf{Ours}  &$\bm{88.79_\pm 0.3}$ & $\bm{{87.60}_{\pm 0.8}}$ & ${.030}_{\pm .001}$ & $\bm{{58.65\pm 1.1}}$ & $\bm{{75.82\pm 1.1}}$ &$\bm{{70.55\pm 1.0}}$ \\ 
    \cline{2-10}
    &\multirow{2}{*}{ATM22} 
    &FA & U-Net~\cite{3dunet}  &${91.29_\pm 0.2}$ & ${89.88}_{ \pm0.5}$ & ${.021}_{\pm .001}$ & ${55.93\pm 0.8}$ & ${80.07\pm 0.5}$ &${{71.68\pm 0.6}}$ \\
    &&SkA & \textbf{Ours}             & ${89.60_\pm 0.1}$ & ${86.82}_{\pm 0.5}$ & ${.021}_{\pm .001}$ & ${52.66\pm 0.4}$ & ${76.64\pm 0.7}$ &${{67.31\pm 0.5}}$ \\ 
        \bottomrule
    \end{tabular}
     }
  
   
\end{table*}

\begin{table*} [t]
\centering
    \small
    \caption{Results of ablation study. $GBI$ denotes Geodesic distance-based buffer inference without Gaussian smoothing. The gray shading indicates our final settings for each group of ablation experiments.}
    \label{tab:ablation}
    \resizebox{0.83\textwidth}{!}
    {
    \begin{tabular}{c ccc| cc| c c c c c c}
        \toprule
        \multicolumn{12}{c}{\textbf{1) Ablation for DBI strategy} ($\mathcal{L} = \mathcal{L}_{pce}$ for default, and $DBI = EBI+G^2BI$)} \\
        \midrule
        &\multicolumn{3}{c}{Setting} & \multicolumn{2}{c}{Params.} & \multicolumn{6}{c}{Metrics} \\
        \cmidrule[0.5pt](rl){2-4}
        \cmidrule[0.5pt](rl){5-6}
        \cmidrule[0.5pt](rl){7-12}
      
        
        & $GBI$& $G^2BI$ & $DBI$ &$\delta1$ &$\delta2$& DSC(\%) $\uparrow$  & TPR(\%) $\uparrow$ & FPR(\%) $\downarrow$ & BD(\%) $\uparrow$ & BD ${^*}(\%)\uparrow$ & TD(\%)$\uparrow$\\
        \midrule 
        \multirow{2}{*}{(a)}
        &  $\checkmark$ &  &  & 0.01 & 0.07 & ${75.91}_{\pm 0.7}$ & ${77.66}_{\pm 1.4}$ &${.280}_{\pm .059}$ & ${24.34}_{\pm 2.2}$ & ${37.39}_{\pm 3.7}$ & ${33.59}_{\pm 2.9}$ \\
        &   &  $\checkmark$ & & 0.01 & 0.07 & ${85.68}_{\pm 0.9}$  & ${78.70}_{\pm 1.3}$ &${.009}_{\pm .002}$ & ${38.69}_{\pm 1.3}$ & ${57.63}_{\pm 1.6}$ & ${50.54}_{\pm 1.5}$\\
        \midrule 
        \multirow{4}{*}{(b)}
        &   &   & $\checkmark$& 0.005 & 0.07& ${83.30}_{\pm 0.4}$ & ${73.24}_{\pm 0.9}$ &${.004}_{\pm .002}$ & ${36.89}_{\pm 5.3}$ & ${55.75}_{\pm 6.9}$ & ${49.08}_{\pm 6.7}$ \\  
        &   &   & $\checkmark$& 0.02 & 0.07 & ${72.25}_{\pm 2.3}$  & ${85.40}_{\pm 1.6}$ & ${.250}_{\pm .047}$ & ${53.74}_{\pm 4.3}$ & ${71.42}_{\pm 3.7}$ & ${65.00}_{\pm 4.1}$\\ 
        &   &   & $\checkmark$& 0.01 & 0.06& ${82.94}_{\pm 0.5}$  & ${74.97}_{\pm 0.3}$
        & ${.020}_{\pm .001}$& ${39.90}_{\pm 0.8}$  & ${59.28}_{\pm 1.0}$ & ${52.48}_{\pm 1.0}$ \\  
        &   &   & $\checkmark$& 0.01 & 0.08 & ${81.20}_{\pm 3.1}$ & ${78.85}_{\pm 2.0}$
        & ${.060}_{\pm .018}$& ${41.62}_{\pm 1.6}$  & ${60.58}_{\pm 2.1}$ & ${53.58}_{\pm 1.8}$\\ 
        \rowcolor{gray!20}
        &   &   & $\checkmark$& 0.01 & 0.07 & ${86.21}_{\pm 0.3}$  &${78.99}_{\pm 0.7}$ &$ {.010}_{\pm .004}$ & ${40.70}_{\pm 2.9}$ & ${60.21}_{\pm 3.3}$ & ${53.06}_{\pm 3.4}$\\
        \midrule 
        
        \multicolumn{12}{c}{\textbf{2) Ablation for GDP framework ($DBI = EBI+G^2BI$ for default)}}\\
        \midrule 
        & $\mathcal{L}_{pce}$ & $\mathcal{L}_{em}$& $\mathcal{L}_{iggd}$& $\lambda_{1}$&$\lambda_{2}$ & DSC(\%) $\uparrow$  & TPR(\%) $\uparrow$ & FPR(\%) $\downarrow$ & BD(\%) $\uparrow$ & BD ${^*}(\%)\uparrow$ & TD(\%)$\uparrow$ \\
        
        \midrule
        \multirow{3}{*}{(c)}
        & $\checkmark$&             &             & 1.5 & 20 
        & ${86.21}_{\pm 0.3}$  & ${78.99}_{\pm 0.7}$ & ${.010}_{\pm .004}$ & ${40.70}_{\pm 2.9}$ & ${60.21}_{\pm 3.3}$ & ${53.06}_{\pm 3.4}$\\ 
        & $\checkmark$& $\checkmark$&             & 1.5 & 20 
        & ${87.62}_{\pm 0.6}$  & ${81.20}_{\pm 0.7}$ & ${.010}_{\pm .001}$ & ${51.64}_{\pm 1.3}$ & ${69.69}_{\pm 1.2}$ & ${63.93}_{\pm 1.4}$\\
        \rowcolor{gray!20}
        & $\checkmark$& $\checkmark$& $\checkmark$ &1.5 & 20 
        &${88.79}_{\pm 0.3}$&${87.60}_{\pm 0.8}$& ${.030}_{\pm .001}$ & ${58.65}_{\pm 1.1}$ & ${75.82}_{\pm 1.1}$& ${70.55}_{\pm 1.0}$ 
        \\
        \midrule
        \multirow{4}{*}{(d)}
        & $\checkmark$& $\checkmark$& $\checkmark$ & 1 & 20 
        & ${87.82}_{\pm 1.1}$  & ${86.50}_{\pm 1.1}$ &${.030}_{\pm.001}$ &  ${58.98}_{\pm 1.4}$ & ${75.39}_{\pm 1.0}$ & ${70.23}_{\pm 1.2}$
        \\
        & $\checkmark$& $\checkmark$& $\checkmark$ & 2 & 20 
        & ${87.50}_{\pm 0.9}$  & ${85.36}_{\pm 2.7}$ & ${.030}_{\pm .001}$ & ${56.04}_{\pm 3.5}$ & ${73.35}_{\pm 3.7}$ & ${67.84}_{\pm 3.5}$
        \\ 
        & $\checkmark$& $\checkmark$& $\checkmark$ & 1.5& 10 
        & ${88.19}_{\pm 0.3}$  & ${85.90}_{\pm 0.8}$ &${.030}_{\pm.001}$ &  ${55.62}_{\pm 1.8}$ & ${72.28}_{\pm 1.5}$ & ${67.27}_{\pm 1.5}$
        \\
        & $\checkmark$& $\checkmark$& $\checkmark$ & 1.5& 40
        & ${88.75}_{\pm 0.4}$  & ${86.77}_{\pm 1.0}$ &${.030}_{\pm.001}$ &  ${56.13}_{\pm 0.8}$ & ${73.81}_{\pm 0.6}$ & ${68.35}_{\pm 0.7}$
        
        \\
        \bottomrule
    \end{tabular}
    }
\end{table*}
        

\noindent{\textbf{Comparative Test.}}
In Table~\ref{tab:sota}, we compare our approach with other sparse-supervised learning methods, including regularization learning~\ucite{grandvalet2004semi,tang2018regularized,zhai2023PA-Seg} and dynamic pseudo-label learning~\ucite{luo2022scribble,lee2020scribble2label}. Note that some of the methods (marked by $\dag$) utilize our mask proposal $\mathcal{MP}_\odot$ during training instead of the SkA, thus avoiding the crash of direct learning.  
Results demonstrate that \textbf{1)} our method achieves superior performance, outperforming the 2nd approach~\cite{grandvalet2004semi} with  considerable margins (+6.4\% TPR, +7.0\%  BD, +6.1\% BD${^*}$, +3.6\% TD). We argue that the regularization learning and dynamic pseudo-label learning approaches emphasize learning a more confident and robust segmentation, which hinders the network from achieving better recall for high-frequency details (such as airway edges or fine peripheral branches). \textbf{2)} Our method achieves segmentation performance close to fully supervised learning with only 1.96\% voxel annotations. \textbf{3)} Additional \textbf{Cross-dataset validation} (i.e., train on BAS, test on ATM22) also confirms the strong generalization ability of our algorithm. 

\noindent{\textbf{Ablation Study.}}
Ablation study is provided in Table~\ref{tab:ablation} to evaluate the importance of the two key designs: 1) label propagation from SkA (i.e., the DBI strategy) and 2) GDP framework. 
From \textbf{1)}, we observe that the initial label propagation strategy and parameter tuning significantly impact the model's performance. Particularly, i) the Gaussian smoothing ($GBI \to G^2BI$) contributes a substantial improvement of 9.77\% in DSC and 16.95\% in TD, primarily attributed to the enhanced consistency within labels, thereby enabling the reliable label propagation. ii) the degree of airway label expansion~(i.e., the selection of $\delta1$) has a greater influence. A smaller extent implies less but more accurate airway label diffusion, while a larger extent implies more but noisier label diffusion, thus facing higher risks of false positives. Additional visual comparisons of various label propagation strategies and parameter selections can be found in the Suppl. Material. Besides, \textbf{2)} the ablation study of GDP confirms that the addition of entropy minimization regularization loss and IGGD prediction loss significantly improved segmentation performance. The former strengthens the model's predictive capability for labeled regions in mask proposals (manifested in better topological-level metrics), but has limited performance improvement in uncertain regions (manifested in limited TPR improvement). IGGD, as soft geometry-aware information, provides the model with more scale and edge perception, resulting in significant performance improvements at both topological and voxel levels (+6.4\% TPR and +7.0\% BD).

\noindent{\textbf{Qualitative Results.}}
Fig.~\ref{result} gives an intuitive performance comparison for the above methods on both a moderate case (a) and a challenging case (b). It can be observed that our method maintains less false positive detection even in noisy images, while detecting more distal branches compared to other methods. Additionally, we achieve comparable performance with the fully supervised method and even superior topological metrics in some cases.

\section{Conclusion}
Tailored to label-effective airway segmentation, we introduce a novel skeleton annotation strategy, which alleviates the annotation burden, preserves complete topology, and enhances annotation consistency and accuracy. Based on that, we further propose a skeleton-supervised learning method. To address the extremely sparse supervision challenge, a dual-stream buffer inference strategy is first proposed to realize initial label propagation. Then, a geometry-aware dual-path propagation learning method is presented in a dual-supervision manner to further achieve complementary label propagation learning. Our proposed method achieved excellent airway segmentation performance with only 1.96\% annotated voxels, approaching the fully supervised performance. This shows great potential for generalization to more tubular structure segmentation tasks.
\bibliographystyle{unsrt}
\bibliography{paper1987}
%




\end{document}